\newcounter{myctr}

\documentclass{article}
\usepackage{url}
\usepackage{epsf}
\usepackage{graphicx}
\usepackage{subfigure}
\usepackage{indentfirst}
\begin{document}

\makeatletter
\def\@biblabel#1{[#1]}
\makeatother

\markboth{Kumar,Rao}{Failover In Cellular Automata}

%
%

\title{FAILOVER IN CELLULAR AUTOMATA}

\author{SHAILESH KUMAR\footnote{shailesh.kumar@iiitb.net} \hspace{1in} SHRISHA RAO\footnote{shrao@ieee.org}}


%

\maketitle

\begin{abstract}

  A cellular automata (CA) configuration is constructed that exhibits
  emergent failover.  The configuration is based on standard Game of
  Life rules.  Gliders and glider-guns form the core messaging
  structure in the configuration.  The blinker is represented as the
  basic computational unit, and it is shown how it can be recreated in
  case of a failure.  Stateless failover using primary-backup
  mechanism is demonstrated.  The details of the CA components used in
  the configuration and its working are described, and a simulation of
  the complete configuration is also presented.

\end{abstract}


\section{Introduction}
   
Cellular automata (CA) have been widely used to model complex systems.
CA succinctly model the self-organizing and emergent properties of
complex systems. Together with their parallelized structure, Cellular
automata make one of most suitable models for complex systems.  Simple
CA constructions often yield collective complex
behaviors~\cite{ca_and_complexity_94}.
   
In addition to modeling complex systems, CA themselves can be
considered as computational units.  Such consideration comes under the
category of non-standard models of computation and is called Collision
Based Computing~\cite{melanie}.  Various logic functions such as NOT,
AND and OR have been constructed in the ``Game of Life'' CA.  The Game
of Life (GOL)~\cite{berlekamp82} is a popular CA which has yielded
many complicated patterns based on its simple set of rules.  Many
complex patterns arise out of some of the basic patterns that occur in
the GOL.  We use some of the basic patterns such as gliders, blinkers
(oscillators), still-life (invariants) and
glider-guns~\cite{berlekamp82} that occur in the GOL to set up our
configuration.  Various circuits can be built up based on the logical
functions that forms one of the techniques to demonstrate universal
computation~\cite{melanie}.  Universal Turing machines have also been
simulated in many different CAs~\cite{Burks70a,Burks70b} and in
particular the GOL also.  Gliders and glider-guns form the key
patterns that are used in the universal Turing machine in the GOL.
From the computing perspective of our construction we represent a
blinker as our basic computational unit. It can be viewed as two-bit
counter and we show how its computation can continue in case of
failure.

In this paper, we present a basic model for failover as a complex
system behavior.  Failover is a widely used distributed-system concept
used to make systems highly available.  Many different techniques of
achieving failover have been discussed~\cite{1228010,1050541}.
Failover is one of the techniques to make a system fault tolerant,
which is a system that continues to operate even in case of certain
faults.  Failover techniques involve the use of redundant
components~\cite{356680,356729}.  
We achieve failover in CA by making use of redundant CA components.
We use the primary-backup approach~\cite{807732,primary-backup} to
implement failover.  In this approach, a standby module takes over the
active module when there is a failure in the active module.  Usually
the active module is called the \emph{primary} and the standby is
called the \emph{backup}.  This ensures that the system as a whole is
available even if there are failures in some of its components.  The
primary and the backup exchange messages between them which are
usually called the \emph{heartbeat} (which may colloquially be
described as ``I'm alive'') messages.  Initially the primary system is
in the active state and performs the required tasks.  It also sends
the heartbeat message to the backup system.  The backup is usually in
a passive state and is not involved in the actual work.  If the
primary module fails then the backup does not receive any heartbeat
messages from the primary, and moves from its standby state and takes
over the work.

Failover is mainly of two types:
\begin{itemize}
\item \emph{Stateless} failover: In this type of failover, the backup
  does not have any state information of the primary and in case of
  failure, it restarts the computation as if newly started.
\item \emph{Stateful} failover: In this type of failover, the backup
  maintains the state information of the primary and in case of a
  failure, it resumes from this state.
\end{itemize}

Our discussion in this paper restricts to stateless failover.
Also, we assume the failure of complete component or module
rather than failure of individual cells.  Our model is based on
fail-stop in which the system as a whole comes to a halt in case of
failures.  We also assume the presence of a global synchronization
clock. 

Section~\ref{related} gives a brief description of related work.  In
Section~\ref{model}, we present the model for a failover in
CA. Section~\ref{components} describes the different components used
in our construction for the failover
configuration. Section~\ref{simulation} gives the details of the
failover construction and the simulation setup and working.  We
finally provide the conclusion in Section~\ref{conclusion}. The
appendix section gives a description about CA and the basic GOL
components used.
    
\section{Related Work} \label{related}
   
Reliable computation with cellular automata involving probabilistic
fault models have been studied in one-dimensional cellular
automata~\cite{808730}.  Self-repairing constructions are used to deal
with these faults.  Synchronous systems, which require the existence of
a fault-free global synchronization clock, are assumed.  Asynchronous
extensions of reliable computation have been done on two-dimensional
cellular automata that perform computation with a probability of
meeting reliability requirements~\cite{async}.  Transient models of
\emph{fault tolerance in CA at high rates} have been studied and fault
rate bounds have also been derived for these
models~\cite{mccann-2007}.
   
Evolutionary algorithms have been applied to cellular automata to
determine the complex global behavior they exhibit or to solve a
particular problem.  Various computational tasks such as density,
synchronization, and random number generation have been demonstrated
on non-uniform Cellular Automata employing genetic algorithms.  In
non-uniform CA, the interaction rules vary in different sections of
the cellular space whereas in uniform CA there are uniform common
rules for the complete cellular space. Fault-tolerant behavior under
random faults have been studied in the non-uniform CA using
evolutionary algorithms to perform computational tasks such as the
density task and synchronization task~\cite{epcm}.

\section{CA Model for Failover} \label{model}
			
In this section, we describe our model for failover in CA (see
Figure~\ref{fig:failover model}).  We use some of the basic GOL
patterns mentioned in the appendix and also some additional patterns.
These form the building blocks for our failover configuration.  To
construct the failover configuration we need:

\begin{itemize}
			\item Primary Module  
			\item Backup Module  
			\item Communication Mechanism 
			\item Trigger for the Failover\\
\end{itemize}

		\begin{figure}[h]
	\centering
		\includegraphics[width=100mm]{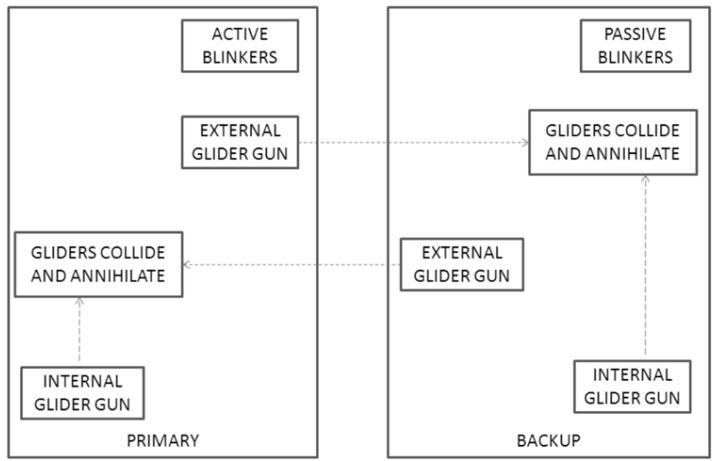}
		\caption{CA model for Failover}
		\label{fig:failover model}
\end{figure}
			
In our configuration, the CA grid is divided into two logical
sections.  One of them forms the primary module and the other the
backup module.  These two sections are made up of similar components
but they differ in their arrangement.  The glider guns and the gliders
together form the communication mechanism between the primary and the
backup.  The gliders also act as the trigger for the backup module in
case of the failure of the primary.
			
Each section has a pair of glider guns and a blinker.  The blinker in
the primary is called the active blinker and in the backup is called a
passive blinker.  One of the glider guns is called the internal glider
gun and the other the external glider gun.  The external glider gun of
the primary communicates with the internal glider gun of the backup
and vice versa.  These essentially form the communication and trigger
mechanism for the system.  The specific details of each component are
mentioned in the following subsection.
			
\subsection{CA components in the model}	\label{components}
			
\textbf{Active Blinker:} The blinker associated with the primary
module is called the active blinker.  It is a configuration where the
cells oscillate between two different states.
						
\begin{figure}[h]
\centering
			\subfigure[State A]{\label{fig:blinker_state_A}\includegraphics[scale=0.5]{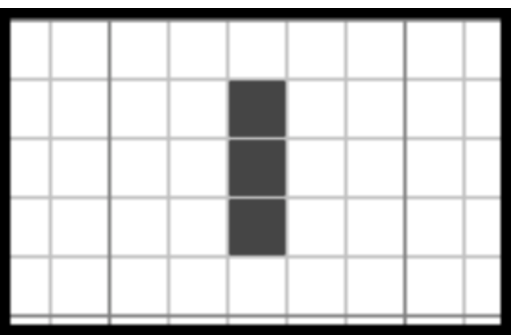}} 
			\subfigure[State B]{\label{fig:blinker_state_B}\includegraphics[scale=0.5]{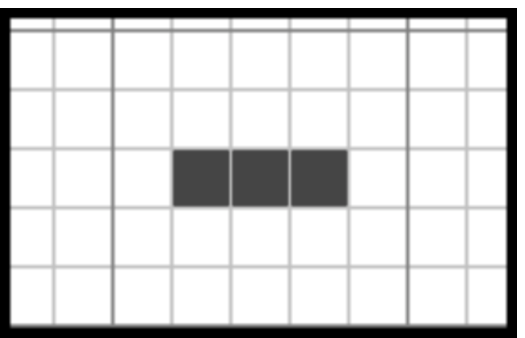}}
			\caption{Blinker}
	\label{fig:blinker}
\end{figure}
\clearpage
			
\begin{figure}[h]
 \centering
 \subfigure[Passive Blinker]{\label{fig:passive_blinker}\includegraphics[scale=0.5]{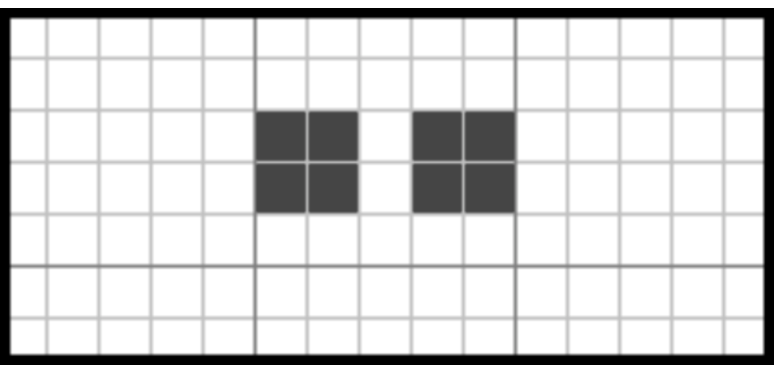}} \\
    \subfigure[]{\label{fig:pb_coll_1}\includegraphics[scale=0.3]{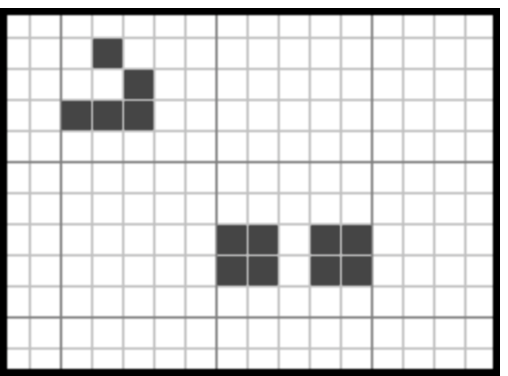}} 
    \subfigure[]{\label{fig:pb_coll_2}\includegraphics[scale=0.3]{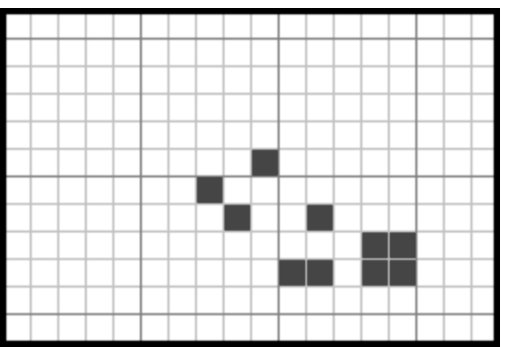}} 
    \subfigure[]{\label{fig:pb_coll_3}\includegraphics[scale=0.3]{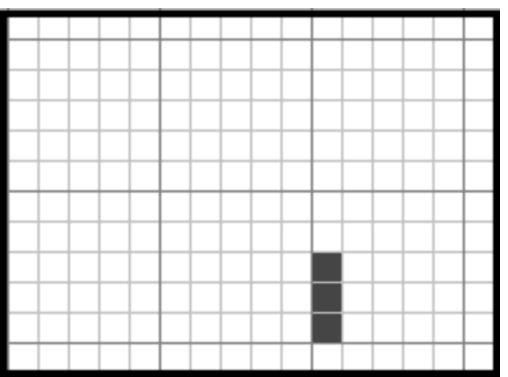}}
    \subfigure[]{\label{fig:pb_coll_4}\includegraphics[scale=0.3]{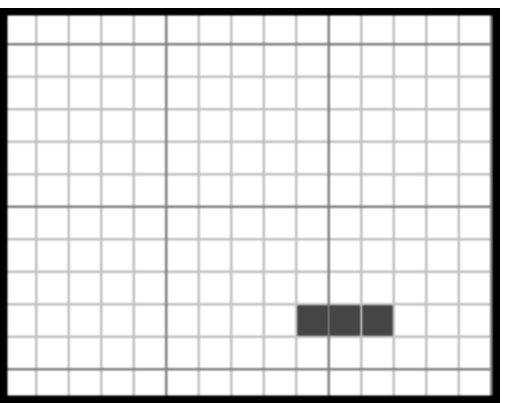}}
   \caption{Collision of Passive Blinker with Glider}
  \label{fig:coll_passive_blinker}
\end{figure}

\textbf{Passive Blinker:} It is a specific still-life configuration
that transforms into a blinker when collided with by a glider.  One of
the passive blinkers we use is as shown in
Figure~\ref{fig:passive_blinker}.

\begin{figure}[h]
 \centering
 \subfigure[]{\label{fig:boat_coll_0}\includegraphics[scale=0.3]{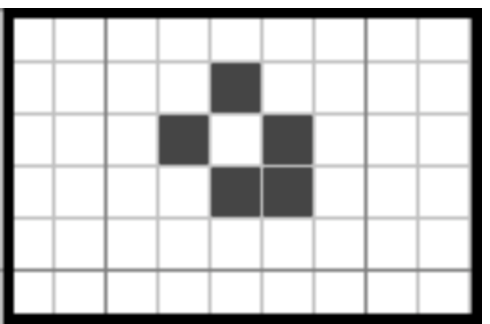}}\\
    \subfigure[]{\label{fig:b_coll_1}\includegraphics[scale=0.3]{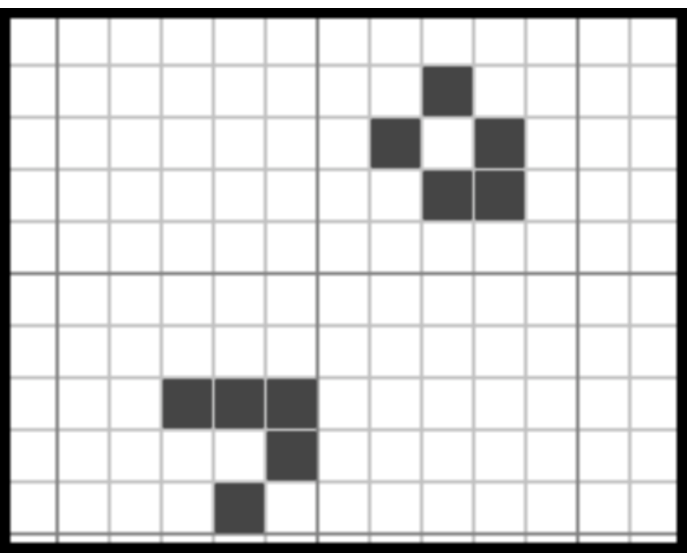}}
    \subfigure[]{\label{fig:b_coll_2}\includegraphics[scale=0.3]{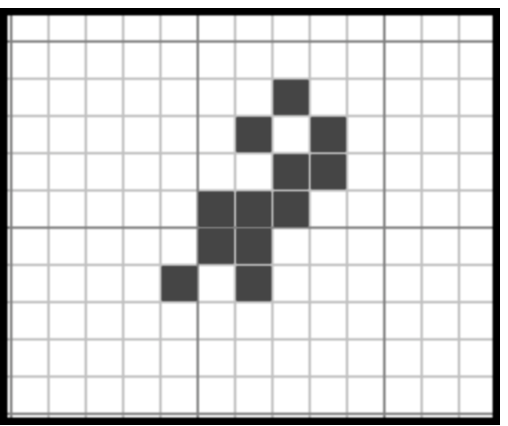}} 
    \subfigure[]{\label{fig:b_coll_3}\includegraphics[scale=0.3]{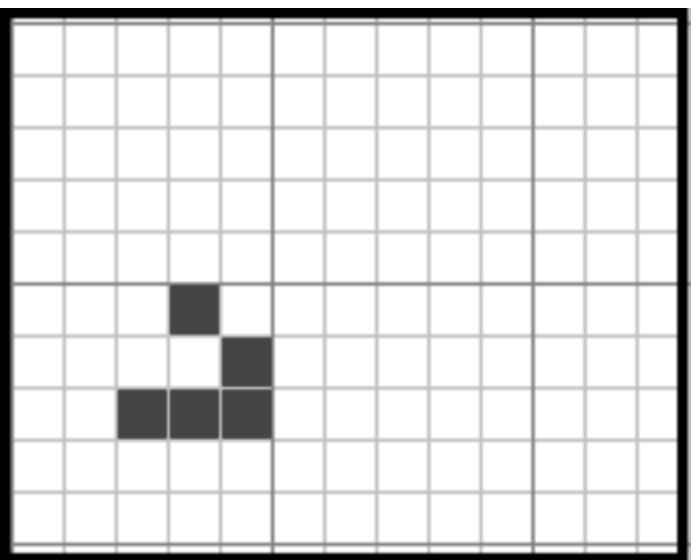}}
     \caption{Glider reflected by $90^{\circ}$}
  \label{fig:boat}
\end{figure}

\textbf{Glider Reflector:} This configuration changes the angle of a
glider by $90^{\circ}$.  This configuration is also called as the
\emph{boat}.  It destroys itself after the glider has been reflected
and so is a one-time reflector.  The working is as shown in
Figure~\ref{fig:boat}.

\clearpage	

\textbf{P-92 Glider Gun}~\cite{p92}: It is similar to the Gosper Gun
(appendix) except that it emits a glider every 92 generations.

\begin{figure}[h]
	\centering
		\includegraphics[width=65mm,height=60mm]{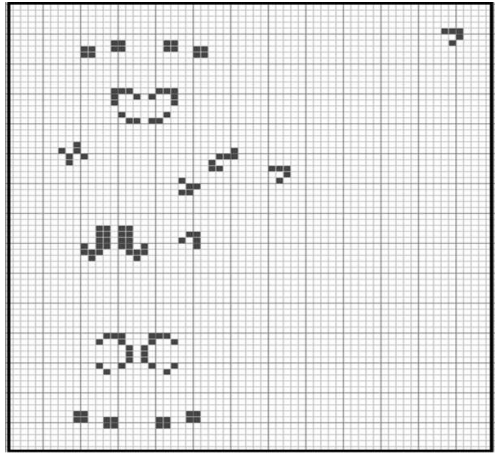}
		\caption{Period-92 Glider Gun}
	\label{fig:p92_glider_gun}
\end{figure}

\begin{figure}[h]
 \centering
    \subfigure[]{\label{fig:glider_coll_1}\includegraphics[scale=0.3]{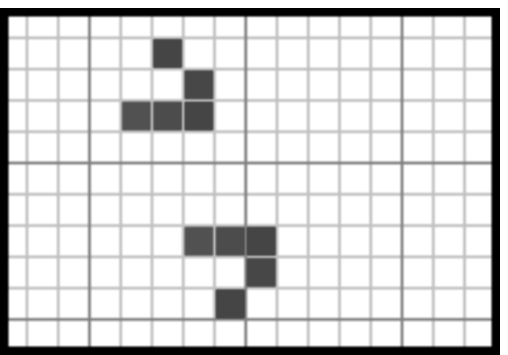}}
    \subfigure[]{\label{fig:glider_coll_2}\includegraphics[scale=0.3]{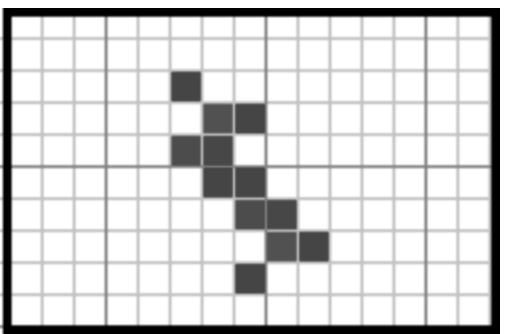}} 
    \subfigure[]{\label{fig:glider_coll_3}\includegraphics[scale=0.3]{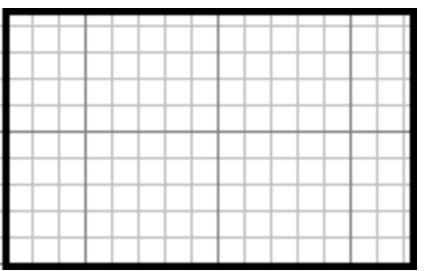}}
     \caption{Annihilating Gliders}
  \label{fig:collision}
\end{figure}

\textbf{Collision between Gliders:} When two gliders collide with each
other at a specific angle, they annihilate.  The working is as shown in
Figure~\ref{fig:collision}.
		
\section{Simulation and Analysis} \label{simulation}

\begin{figure}[h]
	\centering
		\includegraphics[width=60mm,height=40mm]{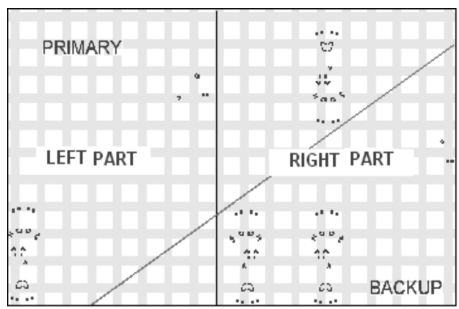}
		\caption{Configuration Figures Interpretation Helper}
	\label{fig:interp_conv}
\end{figure}

\begin{figure}[h]
	\centering
		\subfigure[]{\label{fig:init_setup_left_part}\includegraphics[width=3in,height=2.7in]{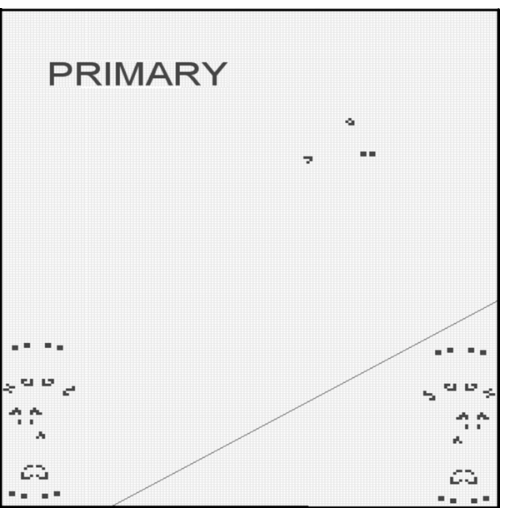}}
		\subfigure[]{\label{fig:init_setup_right_part}\includegraphics[width=3in,height=2.7in]{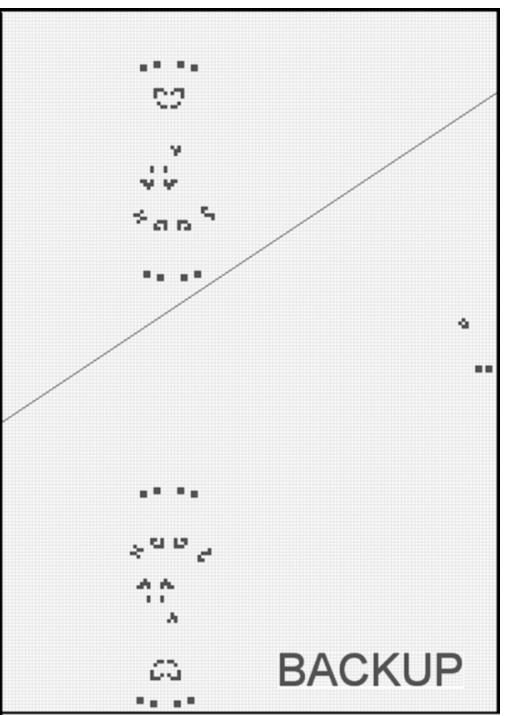}}
		\caption{Failover Configuration}
	\label{fig:init_setup}
\end{figure}

A complete setup for the configuration is as shown in
Figure~\ref{fig:init_setup}.  The dimensions of the grid are
420$\times$200.  The configuration is shown as a pair of two figures,
one representing the left part and the other right part, but the
complete configuration is to be interpreted as a concatenation of
right part to left part as shown in Figure~\ref{fig:interp_conv}.

The grid is divided into two logical sections, namely, the primary and
the backup.  The primary contains a set of two P-92 glider guns 207
cells apart horizontally and 60 cells apart vertically.  One of them
acts as an external glider gun that emits gliders towards the backup
and other an internal glider gun that emits gliders inside the logical
section in the module.  The horizontal distance is measured from the
rightmost active cell of the internal glider gun to the leftmost
active cell of the external glider gun.  Similarly, the vertical
distance is the distance from the topmost active cell of internal
glider gun to the bottom-most active cell of the external glider gun.
The glider reflector is placed at a location such that gliders of the
internal glider gun may collide with it and get reflected by
$90^{\circ}$.  In addition to these, the primary also has a trigger
glider.  It is also placed such that it collides with reflector.  A
passive blinker is placed at a location such that it is at an angle
suitable to be transformed into an active blinker when collided with
by the reflected glider.  The backup contains similar components as
that of the primary but they are slightly different in their
structures.  The backup contains a set of two P-92 glider guns but
these are 39 cells apart horizontally.  In case of backup, the
internal glider gun is translated one cell up compared to the external
glider gun.  It also contains a glider reflector and passive blinker
placed at locations to get a similar effect as that mentioned for the
primary.  One important difference is that there is no input trigger
glider in the backup.

\begin{table}[ht]
\caption{Interaction between different CA components } 
\centering 
\begin{tabular}{c c c} 
\hline\hline 
Component   & Collides with Component\ & Result \\ [0.5ex] 
\hline 
Glider from Backup's  & Glider from Primary's & Annihilation \\ 
Internal Glider Gun &  External Glider Gun&\\ 

Glider from Backup's & -- & Collision with\\
 Internal Glider Gun 	&&Passive Blinker and\\
						&&transformation into\\
							&&Active Blinker\\

Input Glider  & Passive Blinker & Transformation into   \\
&&Active Blinker\\

Glider & Boat & Change the direction  \\
&& of glider by $90^{\circ}$\\
\hline 
\end{tabular}
\label{table:interact_table} 
\end{table}
\clearpage

The simulation has controls such as $KillPrimary, ResetBackup, Init$
that aid in simulating the different conditions of the system.

\begin{itemize}

\item $Init$: This action sets the cells of the primary and backup
  sections to be in their initial configurations.
\item $KillPrimary$: This action clears all the cells in the primary
  section of the grid.  This action is used to bring down the primary
  module, i.e., to simulate a failure in the primary.
\item $ResetBackup$: This action restores the cells of the backup to
  its standby state. The standby state is exactly same as that of the
  initial backup's state.

\end{itemize}

The system initially is brought up in a start state configuration.  In
this state, both blinkers are passive.  When the system is started,
the input trigger glider in the primary collides with glider-reflector
and changes its direction by $90^{\circ}$.  Now it collides with the
passive blinker in the primary and transforms it into an active one.
Meanwhile the communication mechanism of the gliders is also triggered
to start.  The glider from the primary's external glider gun is the
heartbeat message sent to the backup and vice versa.  The gliders from
the external gun move to the backup section and collide with the
gliders of the internal glider gun of the backup.  As long as the
gliders collide and annihilate, the backup may be said to infer that
the primary is alive.  There is a similar heartbeat message coming
from the external glider gun which collides with primary's glider gun,
so that the primary infers that backup is alive.

\begin{figure}[h]
	\centering
\subfigure[]{\label{fig:primary_down_left_part}\includegraphics[width=2.8in,height=2.7in]{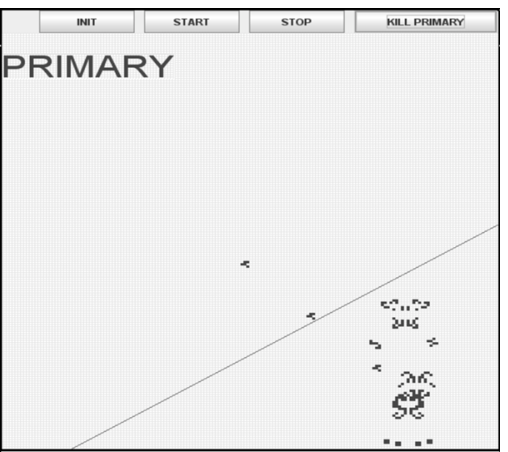}}
\subfigure[]{\label{fig:primary_down_right_part}\includegraphics[width=2.8in,height=2.7in]{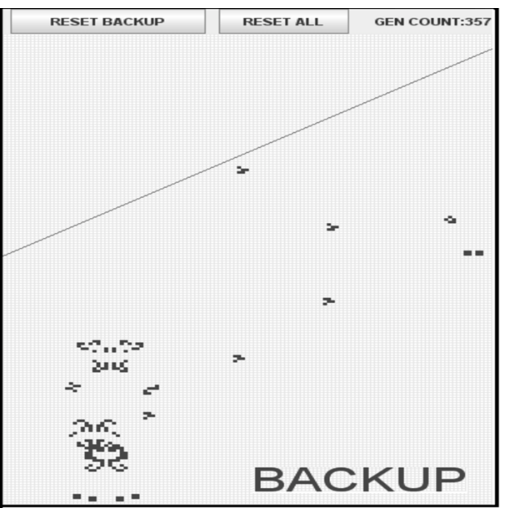}}
\caption{Primary Down and Before Backup becomes Primary}
	\label{fig:primary_down}
\end{figure}

\begin{figure}[h]
	\centering
\subfigure[]{\label{fig:backup_becomes_primary_left_part}\includegraphics[width=3in,height=2.7in]{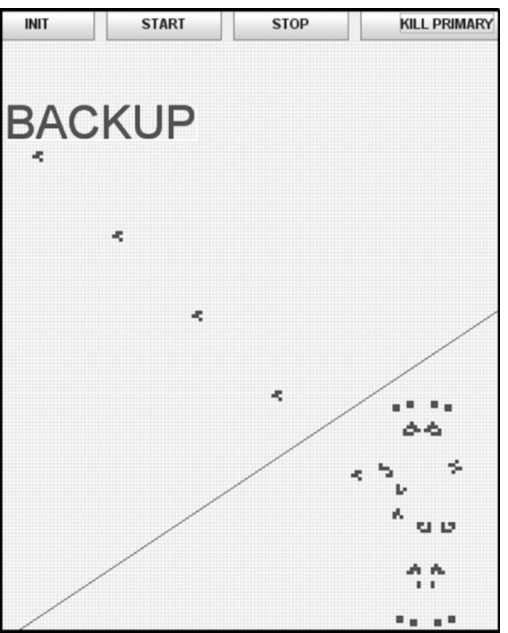}}
\subfigure[]{\label{fig:backup_becomes_primary_right_part}\includegraphics[width=3in,height=2.7in]{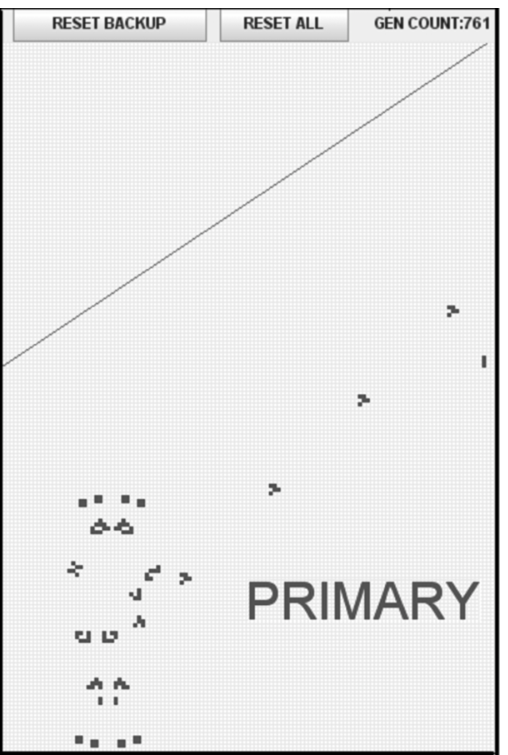}}
		\caption{Primary Down and Backup becomes Primary}
	\label{fig:backup_becomes_primary}
\end{figure}

\textbf{PrimaryDown:} This condition is simulated by invoking the
KillPrimary action. The cells of the primary section are cleared off.
This is similar to bringing down the primary module.  The
configuration is as shown in Figure~\ref{fig:primary_down}.  In this
case, there is no glider emitted from the external gun of the primary.
Therefore the passive blinker in the backup module is triggered by its
internal glider gun and becomes active. Now, the backup becomes the
primary and continues functioning.  The configuration is as shown in
Figure~\ref{fig:backup_becomes_primary}.

\clearpage
	
\textbf{ResetBackup:} The backup is reset by invoking the
$ResetBackup$ action. The cells of the current backup section are set
to the standby state. This action is invoked after primary has gone
down and when the backup module has become the current primary.

\begin{figure}[h]
	\centering
	
			\subfigure[]{\label{fig:backup_reset_left_part}\includegraphics[width=3.1in,height=2.7in]{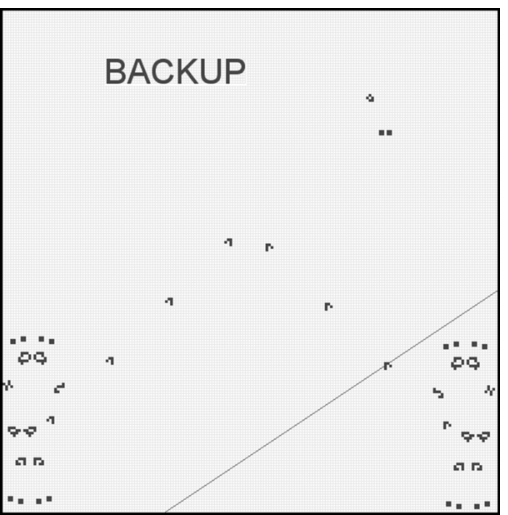}}
		\subfigure[]{\label{fig:backup_reset_right_part}\includegraphics[width=3.1in,height=2.7in]{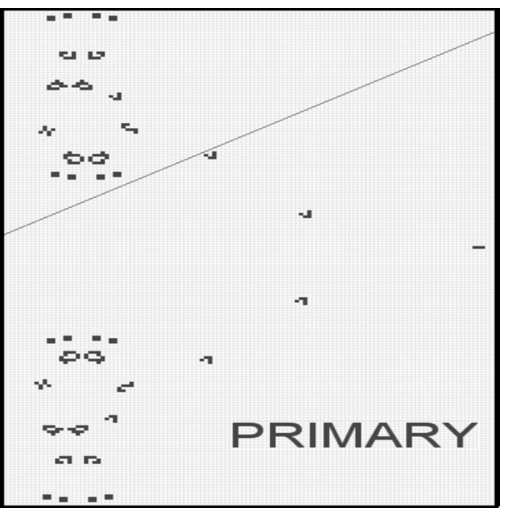}}
		\caption{After Backup is Reset}
	\label{fig:backup_reset}
\end{figure}

When the failed primary module is brought back, it comes up in its
start state which has a passive blinker.  Now, this acts as a backup
and continues the exchange of messages through the glider guns.  The
configuration is as shown in Figure~\ref{fig:backup_reset}.

In the failover configuration discussed here, in case of a primary
failure, the backup waits until all the messages (gliders) that have
been sent by the primary before it went down are received.  The
maximum time for the backup to come into action would be the (number
of gliders present in the communication path times 92 generations) +
number of generations required for the internal glider to collide with
the passive blinker.

When the backup is reset, it is necessary that it is synchronized with
the primary's external glider gun.  If the glider guns are not
synchronized, the gliders may not collide at the appropriate angles
and therefore may not annihilate. The backup needs to be reset at any
\((N \times 92)^{th}\) (where \(N = 1,2,3, \ldots\)) generation for
the communication mechanism to resume properly.

\section{Conclusion} \label{conclusion}
   
In this paper, we have shown that the real world distributed-system
concept of failover can be modeled using cellular automata.  We
focused on stateless failover and constructed a cellular automata
configuration that demonstrates failover using the standard
game-of-life rules.  The period-92 glider guns were used as the basic
communication and trigger mechanism between the primary and backup.
We represent blinkers as our basic computational units and show how
the backup's passive blinker transforms into an active on failure of
the primary module.  We also showed that the backup, when reset, comes
back as a hot standby and whole switching process of primary-backup
can continue indefinitely as long as there is a single (primary)
failure and the backup can be reset. A possible extension of this
model would be to demonstrate a stateful failover in which case the
primary and the backup would maintain state information and in case of
a failure the backup takes over from where the primary module left off
before going down.  This construction can also find its use as a
reusable component in larger complex configurations.



\begin{thebibliography}{10}
\providecommand{\urlprefix}{}
\expandafter\ifx\csname urlstyle\endcsname\relax
  \providecommand{\doi}[1]{doi:\discretionary{}{}{}#1}\else
  \providecommand{\doi}{doi:\discretionary{}{}{}\begingroup
  \urlstyle{rm}\Url}\fi

\bibitem{807732}
Alsberg, P.~A. and Day, J.~D., A principle for resilient sharing of distributed
  resources, in \emph{ICSE '76: Proceedings of the 2nd International Conference
  on Software Engineering} (IEEE Computer Society Press, Los Alamitos, CA, USA,
  1976), pp. 562--570.

\bibitem{berlekamp82}
Berlekamp, E.~R., Conway, J.~H., and Guy, R.~K., \emph{Winning Ways For Your
  Mathematical Plays}, Vol.~2 (Academic Press, ISBN 0-12-091152-3, 1982),
  chapter 25.

\bibitem{primary-backup}
Budhiraja, N., Marzullo, K., Schneider, F.~B., and Toueg, S., The
  primary-backup approach, in \emph{Distributed Systems}, ed. Mullender, S.~J.
  (ACM Press, 1993), pp. 199--216.

\bibitem{Burks70a}
Burks, A.~W., Programming and the theory of automata, in \emph{Essays on
  Cellular Automata, University of Illinois Press, Urbana, Chicago, London},
  ed. Burks, A.~W. (1970).

\bibitem{Burks70b}
Burks, A.~W., Von {N}eumann's self-reproducing automata, in \emph{Essays on
  Cellular Automata}, University of Illinois Press, Urbana, Chicago, London,
  ed. Burks, A.~W. (1970).

\bibitem{1050541}
Burton-Krahn, N., Hot{S}wap - Transparent Server Failover For {L}inux, in
  \emph{{LISA} '02: Proceedings of the 16th USENIX Conference on System
  Administration} (USENIX Association, Berkeley, CA, USA, 2002), pp. 205--212.

\bibitem{356680}
Denning, P.~J., Fault tolerant operating systems, \emph{ACM Comput. Surv.}
  \textbf{8} (1976) 359--389.

\bibitem{p92}
Due, B., Period 92 glider gun (2006), {G}ame of Life News,
  \url{pentadecathlon.com}, March 18th, 2009,
  \url{http://pentadecathlon.com/lifeNews/glider_guns}.

\bibitem{808730}
G\'{a}cs, P., Reliable Computation with Cellular Automata, in \emph{STOC '83:
  Proceedings of the Fifteenth Annual ACM Symposium on Theory of Computing}
  (ACM, New York, NY, USA, 1983), ISBN 0-89791-099-0, pp. 32--41,
  \doi{http://doi.acm.org/10.1145/800061.808730}.

\bibitem{gardner83}
Gardner, M., \emph{Wheels, Life, and Other Mathematical Amusements} (W. H.
  Freeman and Company, New York, 1983), ISBN 0-7167-1589-9.

\bibitem{melanie}
Gramss, T., Bornholdt, S., Gross, M., Mitchell, M., and Pellizzari, T.,
  \emph{Nonstandard Computation} (Wiley-VCH, 1998).

\bibitem{mccann-2007}
McCann, M. and Pippenger, N., \emph{Fault Tolerance in Cellular Automata at High
  Fault Rates} (2007),
  \urlprefix\url{http://www.citebase.org/abstract?id=oai:arXiv.org:0709.0967}.

\bibitem{356729}
Randell, B., Lee, P., and Treleaven, P.~C., Reliability issues in computing
  system design, \emph{ACM Computing Surveys} \textbf{10} (1978) 123--165.

\bibitem{1228010}
Singh, K. and Schulzrinne, H., Failover, load sharing and server architecture
  in {SIP} telephony, \emph{Computer Communications} \textbf{30} (2007)
  927--942.

\bibitem{epcm}
Sipper, M., \emph{Evolution of Parallel Cellular Machines: The Cellular
  Programming Approach} (Springer-Verlag, Heidelberg, 1997).

\bibitem{async}
Wang, W., An asynchronous two-dimensional self-correcting cellular automaton,
  in \emph{Foundations of Computer Science, 1991. Proceedings., 32nd Annual
  Symposium} (IEEE, 1991), ISBN 0-8186-2445-0, pp. 278--285,
  \doi{http://ieeexplore.ieee.org/xpl/freeabs_all.jsp?tp=&arnumber=185379&isnu%
mber=4746}.

\bibitem{ca_and_complexity_94}
Wolfram, S., Complex systems theory, in \emph{Cellular Automata and Complexity:
  Collected Papers,Addison-Wesley} (1994).

\end{thebibliography}

\appendix

\section{Cellular Automata}   

A Cellular Automaton is a discrete model of a collection of cells.
The cells can take any of the states from a set of states. At every
time $t$ (sometimes called generation), the state of each cell is
updated. The state of a cell depends on its current state and the
state of its neighbors.  These constitute the rules for the CA. The
cells are usually rectangular although there are various other
geometric shapes such as hexagonal as well.  The cells can be arranged
in n-dimensional space. The most commonly occurring CA are one, two
and three-dimensional although higher dimensions also have been
explored. We restrict our discussion here to 2-dimensional CA. The
most popular CA so far has been the game of life (GOL) created by John
Conway~\cite{berlekamp82}.

The game of life is a 2-dimensional rectangular array of cells and the
cells can exist in either of the two states i.e. alive or dead.  It is
based on simple rules which can be represented as (new\_life,
over\_population, under\_population).
 
 \begin{itemize} 
 \item A dead cell with exactly `new\_life' number of neighbors alive
   becomes alive in the next generation
 \item An alive cell with greater than `over\_population' number of
   neighbors or lesser than `under\_population' number of neighbors
   dies in the next generation.
 \end{itemize}
 
 Conway's game of life rule is given as (3, 3, 2). Many complex
 patterns have emerged out of this simple rule and many more are being
 discovered. The most commonly occurring patterns are described below
 \\
 Still Life: It is a configuration where all the cells stay alive for
 all generations
\begin{figure}[h]
\centering
  	\includegraphics[scale=0.5]{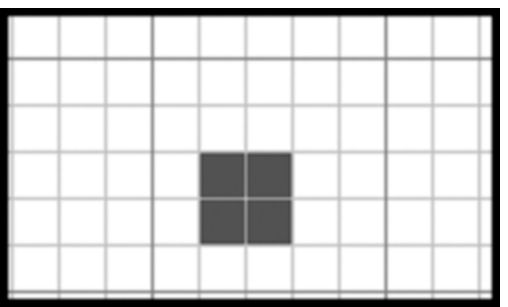}
  	\caption{Still life}
  	\label{fig:still life}
\end{figure}

Blinker/Oscillator: It is a configuration where the cells oscillate
between two different states.
\begin{figure}[h]
\centering
			\subfigure[State A]{\label{fig:apnd_blinker_state_A}\includegraphics[scale=0.5]{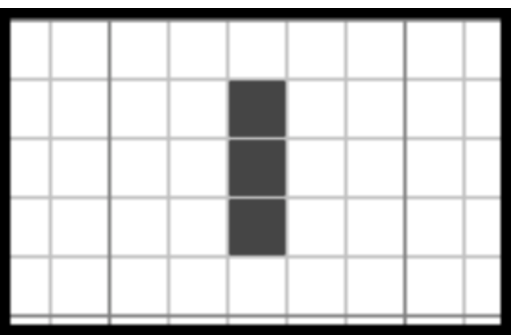}} 
			\subfigure[State B]{\label{fig:apnd_blinker_state_B}\includegraphics[scale=0.5]{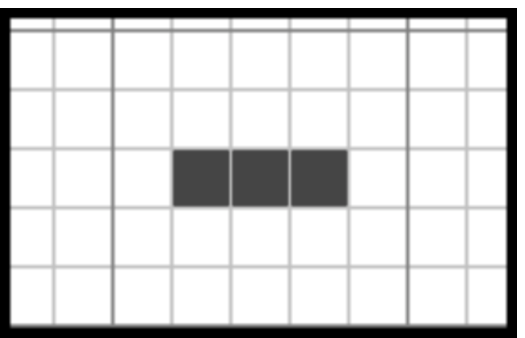}}
			\caption{Blinker}
	\label{fig:apnd_blinker}
\end{figure}

Glider: It is a configuration which translates itself after a certain
number of generations.
\begin{figure}[h]
\centering
		\includegraphics[scale=0.5]{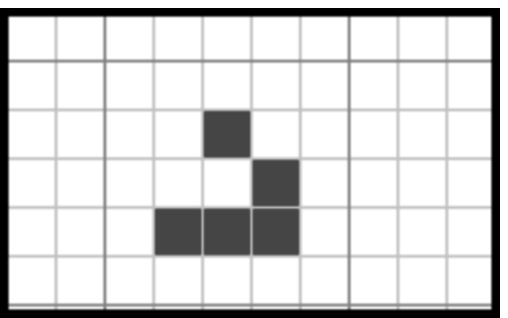}
		\caption{Glider}
	\label{fig:glider}
\end{figure}

Glider gun: It is a configuration that constantly emits gliders. The
first ever glider gun was discovered by Gosper~\cite{gardner83}. It is
a p30 glider gun that emits a glider for every 30 generations.
\begin{figure}[h]
	\centering
		\includegraphics[scale=0.5]{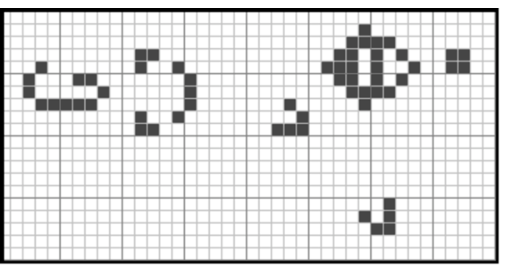}
	\label{fig:gosper gun}
	\caption{Gosper Gun}
\end{figure}

\end{document}